\documentclass[10pt,twocolumn,letterpaper]{article}

\usepackage{paper}

\usepackage[normalem]{ulem}

\definecolor{linkblue}{rgb}{0.21,0.49,0.74}
\usepackage[pagebackref,breaklinks,colorlinks,allcolors=linkblue]{hyperref}

\title{UFMF: Radar-Guided Kinematic Priors for Real-Time 4D Occupancy Forecasting}

\author{%
\begin{tabular}[t]{c}
Zeping Zhang\\
University of Ottawa\\
{\tt\small Zeping.Zhang@uottawa.ca}
\end{tabular}
\hspace{3em}
\begin{tabular}[t]{c}
Zhuoya Zhao\\
University of Ottawa\\
{\tt\small zoezhao0427@gmail.com}
\end{tabular}
\\[1em]
\begin{tabular}[t]{c}
Robert Lagani\`{e}re\\
University of Ottawa\\
{\tt\small laganier@uottawa.ca}
\end{tabular}
}

\begin{document}
\maketitle

\begin{abstract}
Real-time 4D occupancy forecasting requires reliably extrapolating both ego motion and surrounding actor motion under strict latency budgets. Recent occupancy world models improve long-horizon fidelity through generative decoders and 3D volumetric modeling, yet their computational cost makes real-time deployment challenging. Light-weight BEV forecasters meet latency requirements but often blur or suppress fast-moving actors at longer horizons. We present UFMF (Ultra-Fast Multi-modal Forecaster), a real-time forecaster that keeps an attention-free diagonal state-space temporal core while injecting targeted kinematic and geometric structure where lightweight models typically fail. A radar-guided velocity-aware dynamic prior (VADP) decomposes future displacement into a linear extrapolation plus a learned horizon-dependent residual. The resulting per-cell flow fields transport a compact dynamic latent subspace during rollout. Two auxiliary training constraints complement this branch: world-frame dynamic supervision enforces inter-horizon trajectory coherence, and class-conditioned vertical-profile supervision regularizes BEV-to-3D lifting. UFMF operates at 122 FPS on a single RTX 4090 and reaches 0.45 fast-displacement mIoU on nuScenes, outperforming recent occupancy world models at a fraction of their computational cost.
\end{abstract}

\section{Introduction}
\label{sec:intro}
\begin{figure}[t]
\centering
\includegraphics[width=0.85\linewidth]{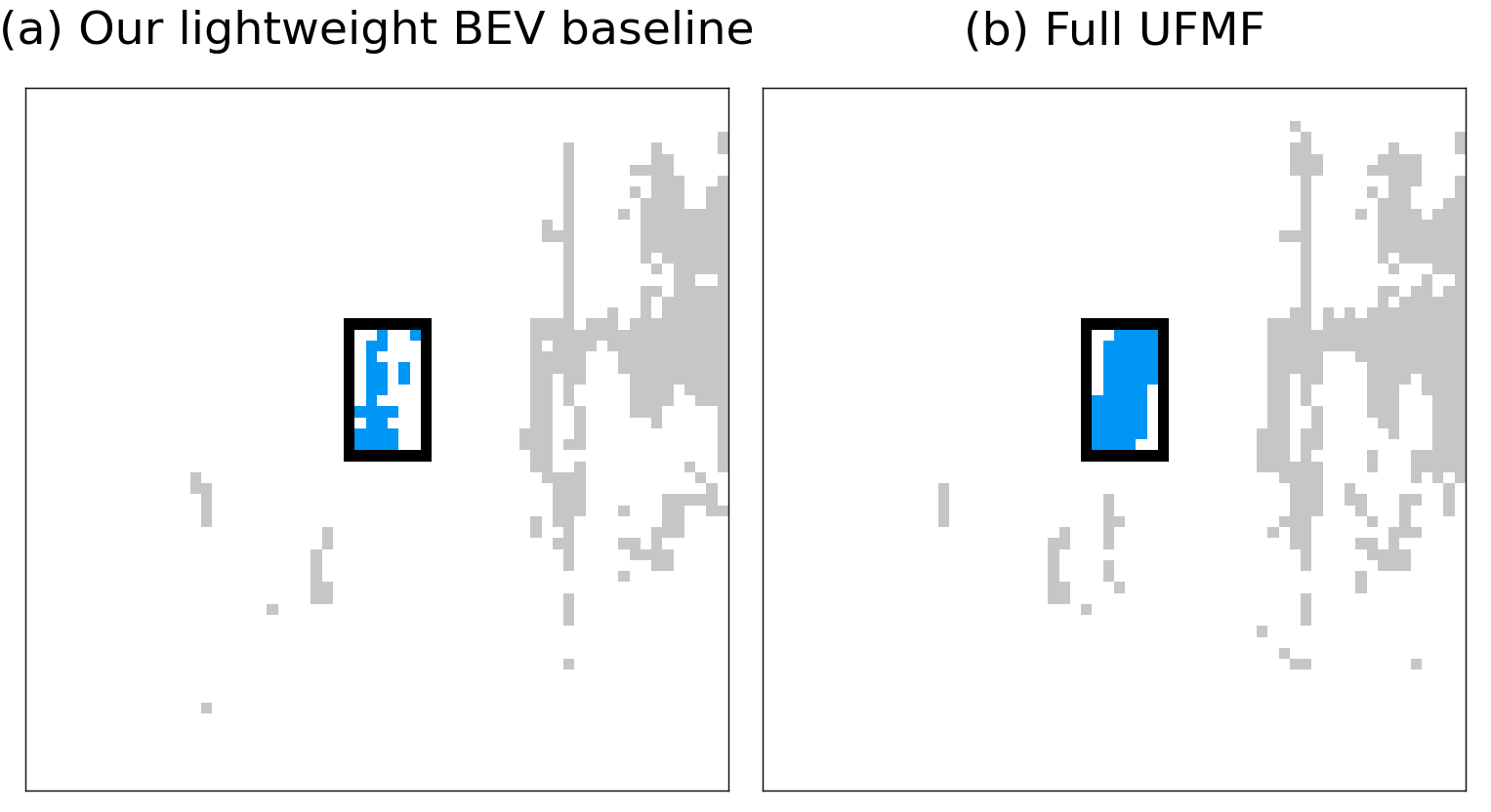}
\caption{Real-data teaser at the last forecast horizon on a nuScenes validation window. (a) Our lightweight BEV baseline (i.e., the first row in \Cref{tab:ablation}: shared backbone without VADP or auxiliary constraints) predicts only a partial future footprint for the boxed fast-moving car. (b) UFMF preserves a fuller, better-aligned footprint.}
\label{fig:teaser}
\end{figure}

Autonomous driving planners rely on predicted occupancy maps to reason about future collisions. When the predicted support of a fast-moving actor diffuses or vanishes at longer horizons, the planner loses the geometry it needs, and the forecast becomes actively harmful rather than merely inaccurate. This problem is most acute in lightweight BEV forecasters: their strict latency budgets leave little room for the heavy spatiotemporal generators that would otherwise stabilize dynamic predictions. \Cref{fig:teaser} illustrates this failure on a real nuScenes~\cite{caesar2020nuscenes} validation window, where a lightweight BEV baseline predicts only a partial future footprint for a fast-moving car, while UFMF preserves a fuller, better-aligned occupancy shape.

Two families of methods address this challenge from opposite directions. Occupancy world models such as OccWorld~\cite{zheng2024occworld}, Drive-OccWorld~\cite{yang2025driveocc}, COME~\cite{shi2025come}, and recent latent-space approaches~\cite{liu2026towards,liu2026occsim} improve long-horizon fidelity through generative 3D decoders and volumetric reasoning, but most run below 25 FPS and the strongest ones operate at only 1--2 FPS, leaving limited headroom for downstream planning. Efficient BEV approaches such as FlashOcc~\cite{yu2023flashocc} maintain real-time speed by performing most computation in 2D and lifting to 3D only at the output via channel-to-height reshaping; however, efficiency alone does not prevent dynamic actors from collapsing when their motion diverges from the ego-warped static background.

UFMF bridges this gap by keeping the temporal core cheap and adding targeted structure only where it matters. Rather than scaling the architecture uniformly, we identify the specific failure mode of lightweight models---dynamic blank-collapse at longer horizons---and address it with three components:
\begin{itemize}
\item A \textbf{Velocity-Aware Dynamic Prior (VADP)} that fuses semantic history deltas with dense radar Doppler cues to predict per-cell displacement fields. The prior decomposes motion into a linear extrapolation plus a learned residual, transporting a compact dynamic latent subspace during the diagonal state-space rollout.
\item \textbf{World-frame dynamic supervision} that evaluates predicted and target occupancy in a common coordinate frame, penalizing trajectories that are locally plausible at each horizon but globally inconsistent over time.
\item \textbf{Class-conditioned vertical-profile supervision} that guides the BEV-to-3D lift head toward semantically plausible height distributions without requiring a heavier 3D decoder.
\end{itemize}
These additions operate alongside the existing diagonal state-space temporal core~\cite{gu2022s4,gupta2022dss} and channel-to-height decoder, preserving a real-time inference path. On nuScenes, UFMF reaches {0.542} overall mIoU and 0.45 fast-displacement mIoU at 122 FPS---outperforming recent occupancy world models by 4.8 percentage points on the fast bucket while running over 100$\times$ faster than the strongest baseline. The ablation trajectory (\Cref{tab:ablation}) further demonstrates that UFMF successfully resolves the dominant bottleneck in lightweight forecasting by injecting explicit kinematic and geometric constraints, rather than simply scaling temporal capacity.

We stress that the contribution is system-level rather than a new atomic operator. Under a strict 8\,ms latency budget, UFMF is the first forecaster to resolve dynamic blank-collapse---a failure mode previously addressable only by heavy generative world models---through targeted kinematic and geometric priors, without scaling the backbone.

\section{Related Work}
\label{sec:related}

\paragraph{BEV forecasting and occupancy world models.}
Bird's-eye-view (BEV) representations have become the standard intermediate for predicting future scenes in autonomous driving. FIERY~\cite{hu2021fiery} introduced probabilistic future instance prediction from surround cameras. StretchBEV~\cite{akan2022stretchbev} extended this with stochastic temporal residuals for longer-horizon diversity, ST-P3~\cite{hu2022stp3} integrated BEV prediction with end-to-end planning, and PowerBEV~\cite{li2023powerbev} showed that parallel multi-scale convolutions can replace autoregressive rollouts. These methods operate in 2D BEV and do not model 3D occupancy evolution.

OccWorld~\cite{zheng2024occworld} pioneered the occupancy world model paradigm by lifting prediction targets to dense 3D voxels and forecasting their evolution with a GPT-like generative transformer over scene tokens. Subsequent works expanded this direction: Drive-OccWorld~\cite{yang2025driveocc} adds motion-conditional normalization and action-conditioned generation, COME~\cite{shi2025come} introduces scene-centric forecasting control, and recent latent-space methods~\cite{liu2026towards,liu2026occsim} scale occupancy world models to longer horizons and larger environments. These methods achieve stronger long-horizon realism through heavy 3D decoders, but the strongest models (Drive-OccWorld, COME) run at only 1--2 FPS, far from real-time deployment. UFMF takes a complementary approach: rather than scaling the decoder, we keep the temporal core lightweight and address the specific failure mode---dynamic blank-collapse---with radar-guided kinematic priors and targeted training constraints. This design recovers most of the dynamic fidelity gap at over 100$\times$ lower latency.

\paragraph{Radar for dynamic perception.}
Automotive radar provides direct radial-velocity measurements via the Doppler effect, a property absent from cameras and LiDAR. RadarNet~\cite{yang2020radarnet} pioneered radar--LiDAR fusion for velocity estimation, CenterFusion~\cite{nabati2021centerfusion} associated radar returns with image detections via frustum projection, and CRN~\cite{kim2023crn} achieved near-LiDAR detection accuracy by using sparse radar as depth anchors for camera BEV transformation. These works exploit radar for perception tasks. UFMF instead uses radar-derived motion cues as a conditioning signal for temporal forecasting. While semantic history can in principle encode motion patterns, doing so requires expensive cross-attention or larger models; radar provides the same velocity evidence directly, enabling lightweight architectures to achieve dynamic fidelity that would otherwise demand significantly more computation.

\paragraph{State-space models for temporal sequences.}
Structured state-space models (SSMs) offer linear-time sequence processing with stable long-range memory. S4~\cite{gu2022s4} introduced the structured parameterization, and DSS~\cite{gupta2022dss} demonstrated that diagonal variants match the full model while being simpler to implement. UFMF adopts the diagonal SSM variant because its per-pixel recurrence is embarrassingly parallel across the BEV grid, preserving real-time throughput while providing stable latent memory across forecast horizons.

\paragraph{Efficient BEV-to-3D lifting.}
FlashOcc~\cite{yu2023flashocc} proposed channel-to-height transformation to avoid expensive 3D convolutions: a 2D network predicts $C \times Z$ channels that are reshaped into voxel logits, enabling occupancy prediction at minimal overhead. BEVFormer~\cite{li2022bevformer} lifts multi-camera features via spatiotemporal deformable attention but targets perception rather than forecasting. UFMF follows the channel-to-height paradigm for efficiency but adds class-conditioned vertical-profile supervision that guides the lift head toward semantically plausible height distributions, addressing the inherent ambiguity of projecting 2D features to 3D without requiring a heavier decoder.

\section{UFMF}
\label{sec:method}

\subsection{Problem setup and model overview}
Given an ego-aligned semantic BEV history $\{S_t\}_{t=1}^{T}$ and rasterized radar Doppler maps $\{D_t\}_{t=1}^{T}$, UFMF predicts future BEV logits
\begin{equation}
Y \in \mathbb{R}^{F \times C \times H \times W}
\end{equation}
and lifted voxel logits
\begin{equation}
Z \in \mathbb{R}^{F \times C \times D \times H \times W},
\end{equation}
where $T$ and $F$ denote the numbers of history and future steps, $H \times W$ is the BEV grid size, $C$ is the semantic vocabulary, and $D$ is the number of height bins. Future ego poses $\{\Pi_k\}_{k=1}^{F}$ are used exclusively during training to warp predictions into a shared world frame for auxiliary supervision (\Cref{sec:worldframe}); they are never accessed at inference.

\begin{figure*}[t]
\centering
\includegraphics[width=\linewidth, trim=200 500 1200 200, clip]{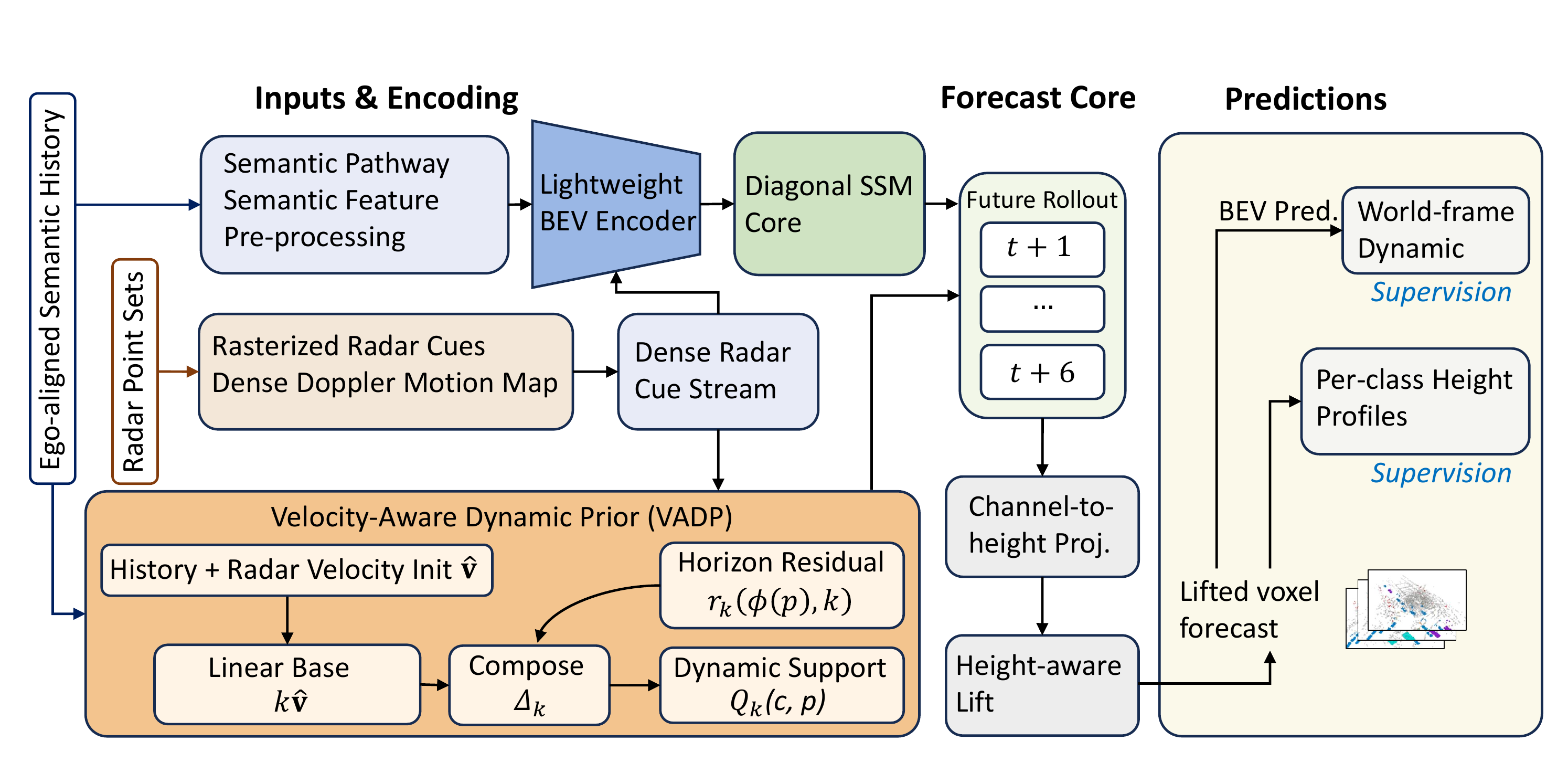}
\caption{Overview of UFMF. Semantic history and rasterized radar cues are encoded by a lightweight BEV encoder and diagonal SSM history encoder, then propagated by a future rollout. VADP uses semantic deltas and radar cues to predict $\Delta_k = k\hat{\mathbf{v}} + r_k$ and dynamic support $Q_k(c,p)$, which guide the rollout state before decoding. The prediction output panel shows the lifted voxel forecast, its BEV projection, and the training-only world-frame and height-profile supervision heads.}
\label{fig:overview}
\end{figure*}

The forecasting pipeline proceeds as follows (\Cref{fig:overview}). A lightweight BEV encoder fuses semantic history with a dense radar cue stream. A diagonal state-space model (SSM)~\cite{gu2022s4,gupta2022dss} encodes the observed window into a latent state. The velocity-aware dynamic prior (VADP) predicts horizon-wise displacement fields from semantic deltas and radar cues, and these fields transport a compact dynamic latent subspace during rollout. Finally, a channel-to-height projection maps the rollout features into voxel logits. The rollout states are internal latent features; the BEV map used for supervision is obtained by projecting the lifted voxel logits back to BEV.

Radar serves two roles: (1) the encoded cue stream is fused with the semantic stream before the SSM encoder and rollout path, and (2) the same stream together with semantic history deltas conditions VADP. The central design question is not how to increase temporal capacity, but how to prevent a lightweight model from collapsing under dynamic uncertainty.

\subsection{Velocity-Aware Dynamic Prior}
\label{sec:vadp}
The dominant source of failure in lightweight BEV forecasting is that dynamic support is predicted too weakly at future horizons. A static ego-warp prior can preserve background structure, but it is incorrect for independently moving actors. UFMF therefore adds an auxiliary dynamic prior only for semantic classes with non-negligible motion, such as cars, trucks, buses, bicycles, motorcycles, and pedestrians. The base predictor still outputs all classes, including background and static structure; the auxiliary prior only adds extra support for dynamic classes.

Radar enters this component through the same dense cue stream used by the Lightweight BEV encoder and the SSM forecasting path in \Cref{fig:overview}. For each BEV cell $p$, we encode the ego-aligned radar history into a per-cell feature map and define $\mathbf{m}(p)$ as the concatenation of three local descriptors at cell $p$: the latest cue, the temporal mean cue, and the recent cue trend. Although raw radar returns are sparse at sensor level, $g_{\text{vel}}$ operates on this rasterized dense BEV cue representation. The sparse-to-dense conversion is a parameter-free grid pooling: for every BEV cell we average the radial Doppler of all radar returns that fall inside that cell, and cells receiving no return remain zero. The main forecasting path uses the same radar stream for temporal conditioning, while VADP reuses the radar-derived motion summary specifically for dynamic extrapolation.

Let $e_t(p)$ denote the shared class-embedding lookup at BEV cell $p$ in history frame $t$, evaluated on the final two ego-aligned semantic-history frames. We estimate a local dynamic velocity field
\begin{equation}
\hat{\mathbf{v}}(p) = g_{\text{vel}}\big([e_T(p) - e_{T-1}(p)\, ;\, \mathbf{m}(p)]\big),
\end{equation}
where $g_{\text{vel}}$ is a lightweight convolutional predictor driven jointly by semantic motion differences and direct radar-derived motion cues. For horizon $k$, a purely linear extrapolation would move the dynamic support by $k\hat{\mathbf{v}}(p)$. We instead decompose future displacement into a linear term plus a horizon-conditioned residual:
\begin{equation}
\Delta_k(p) = k\hat{\mathbf{v}}(p) + r_k\big(\phi(p),\, k\big),
\label{eq:vadp}
\end{equation}
where $r_k(\cdot)$ is a horizon-conditioned residual displacement head, $\phi(p)$ is a local motion context assembled from semantic deltas, radar cues, and the current velocity estimate, and $k$ indexes the forecast horizon via a learned embedding. The residual term is initialized near zero, so the prior starts as linear extrapolation and only deviates when the local evidence supports braking or turning.

We apply the resulting flow field to the one-hot dynamic-class map from the most recent history frame:
\begin{equation}
Q_k(c, p) = \mathcal{W}\big(Q_T(c, p), \Delta_k(p)\big), \qquad c \in \mathcal{C}_{\text{dyn}},
\end{equation}
where $Q_T(c,p)$ is the one-hot indicator of dynamic class $c$ at cell $p$ in the most recent history frame, $Q_k(c,p)$ is the resulting horizon-$k$ prior, $\mathcal{W}(\cdot)$ is bilinear BEV warping, and $\mathcal{C}_{\text{dyn}}$ is the dynamic class set. The warped map $Q_k$ defines a dynamic-class BEV prior and an interpretable support field. We convert the absolute displacements into incremental transport fields
\begin{equation}
\delta_k(p) = \Delta_k(p) - \Delta_{k-1}(p), \qquad \Delta_0(p)=0.
\end{equation}
Let $h_{k-1}(p) \in \mathbb{R}^{C_h}$ denote the rollout state at cell $p$ before horizon step $k$. We form a reduced-channel dynamic branch via a learned $1\times1$ projection $P_{\text{dyn}}: \mathbb{R}^{C_h}\!\to\!\mathbb{R}^{C_d}$:
\begin{equation}
z_{k-1}(p) = P_{\text{dyn}}\big(h_{k-1}(p)\big), \qquad C_d \ll C_h,
\end{equation}
warp that branch by the incremental field,
\begin{equation}
\widetilde{z}_k = \mathcal{W}(z_{k-1}, \delta_k),
\end{equation}
and derive a scalar support gate from the warped prior,
\begin{equation}
\alpha_k(p) = \max_{c \in \mathcal{C}_{\text{dyn}}} Q_k(c,p).
\end{equation}
The transported residual and reinjected state are then
\begin{equation}
r_k^{\text{dyn}}(p) = \alpha_k(p)\big(\widetilde{z}_k(p) - z_{k-1}(p)\big),
\end{equation}
\begin{equation}
h_{k-1}^{+}(p) = h_{k-1}(p) + U_{\text{dyn}}\big(r_k^{\text{dyn}}(p)\big),
\end{equation}
where $U_{\text{dyn}}$ is a learned $1\times1$ projection back to the full hidden dimension. The diagonal SSM rollout is then applied to $h_{k-1}^{+}$ together with the usual horizon drive. This transport branch provides a motion-aligned latent residual while keeping inference fast because only the reduced-channel dynamic branch is warped at each horizon.

No direct velocity labels are used. Instead, $g_{\text{vel}}$ and the residual head are trained end-to-end through the forecasting objective, because $\Delta_k$ determines both the support map $Q_k$ and the transport field $\delta_k$ entering the rollout. In particular,
\begin{equation}
\frac{\partial \mathcal{L}}{\partial \hat{\mathbf{v}}}
= \sum_{k=1}^{F}
\frac{\partial \mathcal{L}}{\partial \Delta_k}
\frac{\partial \Delta_k}{\partial \hat{\mathbf{v}}},
\end{equation}
where $\partial \mathcal{L}/\partial \Delta_k$ accumulates gradients through both the warped support map $Q_k$ and the transported branch $\widetilde{z}_k$. \Cref{fig:overview} summarizes this branch at the right level: encoded radar cues directly condition the velocity initialization and residual refinement, the prior predicts $\Delta_k = k\hat{\mathbf{v}} + r_k$, and the resulting support-guided transport modifies the rollout state before the channel-to-height decoding that produces the final voxel predictions.

\subsection{World-frame dynamic supervision}
\label{sec:worldframe}
Per-horizon overlap in the ego frame alone does not enforce temporally coherent motion. A prediction can achieve reasonable IoU at each individual horizon while producing a trajectory whose support jitters or shrinks inconsistently over time. We therefore evaluate dynamic occupancy in a shared world-aligned frame.

Let $D_k^{\text{pred}}$ and $D_k^{\text{gt}}$ denote the predicted and target dynamic occupancy maps for horizon $k$. Using the known future ego pose, we warp both maps into a shared world frame:
\begin{equation}
\widetilde{D}_k^{\text{pred}} = \mathcal{T}_{k \rightarrow w}(D_k^{\text{pred}}), \qquad
\widetilde{D}_k^{\text{gt}} = \mathcal{T}_{k \rightarrow w}(D_k^{\text{gt}}).
\end{equation}
Here $\mathcal{T}_{k \rightarrow w}(\cdot)$ denotes differentiable warping from the ego frame at horizon $k$ into a shared world frame.
Operating in this common coordinate system is essential: only after removing ego-motion can the pixel-wise difference between consecutive horizons reflect the true physical motion of actors. We apply three complementary terms. The first is a per-horizon dense consistency term on motion-sensitive edges, ensuring that actor boundaries remain sharp in the world frame:
\begin{equation}
\mathcal{L}_{\text{edge}} = \sum_{k=1}^{F} \rho\!\left(\nabla \widetilde{D}_k^{\text{pred}} - \nabla \widetilde{D}_k^{\text{gt}}\right),
\end{equation}
where $\nabla$ is the spatial gradient operator and $\rho(\cdot)$ is a smooth $\ell_1$ penalty.

The second term couples adjacent horizons by penalizing inconsistencies in the inter-frame occupancy change:
\begin{equation}
\mathcal{L}_{\text{td}} = \sum_{k=1}^{F-1} \rho\!\Big(\big(\widetilde{D}_{k+1}^{\text{pred}} - \widetilde{D}_k^{\text{pred}}\big) - \big(\widetilde{D}_{k+1}^{\text{gt}} - \widetilde{D}_k^{\text{gt}}\big)\Big).
\end{equation}
This temporal-difference term cannot be computed in the ego frame because ego rotation and acceleration create spurious pixel displacements; the world-frame transform $\mathcal{T}_{k \rightarrow w}$ eliminates these, leaving only the genuine actor motion field.
The combined world-frame loss is $\mathcal{L}_{\text{wf}} = \mathcal{L}_{\text{edge}} + \beta\,\mathcal{L}_{\text{td}}$, where $\beta$ is a scalar balancing coefficient.

To constrain temporal coherence at the object level, we also penalize the inter-horizon motion of dynamic centroids. To avoid cancellation when multiple actors move in opposite directions, we partition the BEV plane into $M$ non-overlapping spatial zones $\{\Omega_m\}_{m=1}^{M}$ and compute a local centroid within each zone that contains dynamic ground-truth occupancy:
\begin{equation}
\mathbf{c}_{k,m}(\widetilde{D}) = \frac{\sum_{p\,\in\,\Omega_m} p\,\widetilde{D}_k(p)}{\sum_{p\,\in\,\Omega_m} \widetilde{D}_k(p)\,+\,\epsilon}
\end{equation}
The trajectory-shape loss compares centroid displacements between consecutive horizons across all active zones:
\begin{equation}
\mathcal{L}_{\text{traj}} = \sum_{k=1}^{F-1} \sum_{m=1}^{M} w_{k,m}\,
\rho\!\left(\big(\mathbf{c}_{k+1,m}^{\text{pred}} - \mathbf{c}_{k,m}^{\text{pred}}\big) - \big(\mathbf{c}_{k+1,m}^{\text{gt}} - \mathbf{c}_{k,m}^{\text{gt}}\big)\right),
\label{eq:trajloss}
\end{equation}
where $w_{k,m}$ is a non-negative scalar that increases with the magnitude of ground-truth motion inside zone $\Omega_m$, and $\epsilon$ is a small constant for numerical stability. \Cref{eq:trajloss} is critical for fast-moving actors: it penalizes motion that is locally plausible at each horizon but globally inconsistent as a trajectory. Because each zone maintains an independent centroid, actors moving in opposite directions no longer cancel each other out.

\subsection{Class-conditioned vertical-profile supervision}
\label{sec:height}
Efficient occupancy prediction typically performs most reasoning in BEV and lifts the result to 3D only at the end~\cite{yu2023flashocc}. This design is attractive for deployment, but it leaves the vertical structure underconstrained: a correct BEV footprint does not uniquely determine whether the object occupies one voxel near the ground, a tall vertical column, or a semantically implausible height distribution.

UFMF addresses this ambiguity with class-conditioned vertical profiles. Let $P_k^c(z, x, y)$ be the softmax probability that voxel $(z, x, y)$ at horizon $k$ belongs to class $c$. For the BEV cells assigned to class $c$, denoted by $\Omega_k^c$, we define the predicted vertical profile
\begin{equation}
h_k^c(z) = \frac{1}{|\Omega_k^c|} \sum_{(x,y) \in \Omega_k^c} P_k^c(z, x, y).
\end{equation}
Using proxy voxel occupancy supervision $O_k(z, x, y) \in \{0,1\}$, we form the target profile
\begin{equation}
\bar{h}_k^c(z) = \frac{1}{|\Omega_k^c|} \sum_{(x,y) \in \Omega_k^c} O_k(z, x, y).
\end{equation}
The class-height loss is then
\begin{equation}
\mathcal{L}_{\text{height}} = \sum_{k=1}^{F} \sum_{c \in \mathcal{C}_{\text{obj}}}
\rho\!\left(h_k^c - \bar{h}_k^c\right),
\end{equation}
where $\mathcal{C}_{\text{obj}}$ denotes the non-empty, non-background semantic classes. This supervision does not require a heavy recurrent 3D decoder. Instead, it tells the lightweight lift head what vertical support each semantic class should carry.

\begin{figure}[t]
\centering
\includegraphics[width=\linewidth, trim=50 450 1635 420, clip]{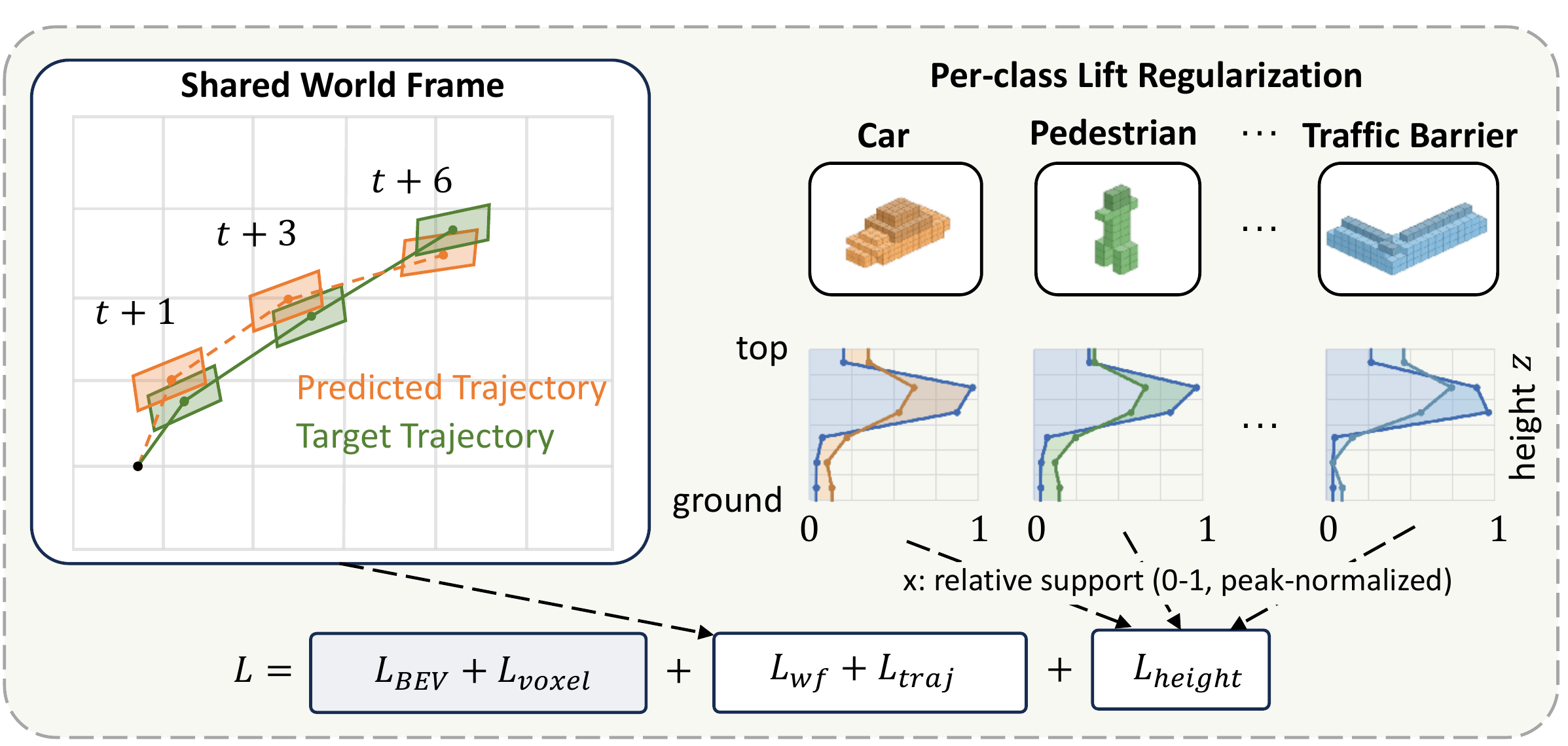}
\caption{Training-only geometric constraints. Left: predicted and target dynamic occupancy are warped into a shared world frame to supervise occupancy consistency and inter-horizon trajectory shape. Right: class-conditioned vertical profiles regularize the BEV-to-3D lift. The footer shows the full training objective.}
\label{fig:training_constraints}
\end{figure}

\subsection{Training objective and inference cost}
\label{sec:objective}
The final objective combines the base BEV segmentation loss with auxiliary geometric constraints:
\begin{equation}
\mathcal{L} = \mathcal{L}_{\text{BEV}}
+ \lambda_{\text{voxel}} \mathcal{L}_{\text{voxel}}
+ \lambda_{\text{wf}} \mathcal{L}_{\text{wf}}
+ \lambda_{\text{traj}} \mathcal{L}_{\text{traj}}
+ \lambda_{\text{height}} \mathcal{L}_{\text{height}}.
\label{eq:objective}
\end{equation}
Here $\mathcal{L}_{\text{BEV}}$ is the semantic BEV forecasting loss, $\mathcal{L}_{\text{voxel}}$ is a binary voxel-occupancy loss derived from proxy occupancy targets, and the $\lambda$ terms are scalar loss weights. The world-frame ($\mathcal{L}_{\text{wf}}$, $\mathcal{L}_{\text{traj}}$) and height-profile ($\mathcal{L}_{\text{height}}$) terms act during training only. At inference, only the transport branch adds computation, and it warps only a compact dynamic latent subspace rather than the full hidden state, preserving the 122 FPS deployment point. Concretely, the branch warps a $C_d{=}32$ channel subspace of the $C_h{=}128$ dimensional hidden state with a single bilinear grid sample per horizon step, adding less than 0.5\,ms in total---far cheaper than the 3D attention or generative decoding used by occupancy world models.

\section{Experiments}
\label{sec:experiments}

\subsection{Experimental setup}
We evaluate on the nuScenes v1.0-trainval split~\cite{caesar2020nuscenes}. Each sample uses 6 ego-aligned semantic BEV history frames and predicts the next 6 future horizons at 2 Hz, corresponding to forecasts up to 3 s.

The voxel grid spans $[-40,40]$ m in both BEV axes with a voxel size of $0.4\times0.4$ m, and uses 20 height bins over $[-5,3]$ m. Targets are sensor-grounded voxel occupancies derived directly from nuScenes LiDAR sweeps rather than dense completion labels. This matches the deployment goal of forecasting measurable future occupancy under a strict runtime budget.

We report overall semantic mIoU over all classes, including background, and motion-stratified semantic mIoU under three actor-displacement buckets: static ($<2$ m), slow ($2$--$10$ m), and fast ($\ge 10$ m). These buckets are defined by target-actor movement over the forecast interval. The bucketed scores reveal whether a model preserves actor support as motion magnitude increases, while mIoU(all) summarizes global semantic fidelity.

Following the standard evaluation protocol for 4D occupancy forecasting, the BEV histories fed to every method are dataset ground truth. OccWorld, Drive-OccWorld and COME are all evaluated this way so that forecasting quality is isolated from upstream perception error, and our setting is strictly aligned with theirs. \Cref{apdx:predhistory} additionally reports UFMF driven by predicted, non-ground-truth histories.

\subsection{Main comparison}
\Cref{tab:main_results} summarizes the main comparison against recent occupancy world models. UFMF achieves the best score on every reported metric, reaching 0.542 mIoU(all) and 0.45 fast-displacement mIoU while running at 122 FPS. Relative to COME, the strongest baseline in this comparison, UFMF improves overall mIoU by 1.5 percentage points (0.542 vs. 0.527) and fast-displacement mIoU by 4.8 percentage points (0.45 vs. 0.402). The margin is larger against OccWorld and Drive-OccWorld, which indicates that the proposed kinematic structure is most valuable when long-horizon actor motion becomes difficult to extrapolate reliably.

The 1.5-point gap in mIoU(all) understates the practical difference, because background and static voxels dominate that average. On the fast-displacement bucket that the method actually targets, UFMF gains 4.8 points over COME and 15.2 points over the lightweight BEV baseline. For downstream planning, preserving the future footprint of a fast-moving vehicle is worth considerably more than a marginal improvement on static structure.

\begin{table*}[t]
\centering
\scriptsize
\caption{Main comparison on a nuScenes~\cite{caesar2020nuscenes} validation set. mIoU(all) averages over all semantic classes, including background. Static/Slow/Fast denote target-actor displacement buckets of $<2$ m, $2$--$10$ m, and $\ge 10$ m, respectively.}
\begin{tabular*}{\textwidth}{@{\extracolsep{\fill}} l c c c c c @{}}
\toprule
Method & mIoU(all) $\uparrow$ & Static $\uparrow$ & Slow $\uparrow$ & Fast $\uparrow$ & FPS $\uparrow$ / ms $\downarrow$ \\
\midrule
OccWorld~\cite{zheng2024occworld} & 0.478 & 0.571 & 0.472 & 0.336 & 23 / 44 \\
Drive-OccWorld~\cite{yang2025driveocc} & 0.509 & 0.601 & 0.506 & 0.381 & $\sim$2 / 647 \\
COME~\cite{shi2025come} & 0.527 & 0.618 & 0.526 & 0.402 & $\sim$1 / 1157 \\
UFMF (ours) & \textbf{0.542} & \textbf{0.632} & \textbf{0.541} & \textbf{0.450} & \textbf{122 / 8} \\
\bottomrule
\end{tabular*}
\label{tab:main_results}
\end{table*}

Occupancy world models are not the only relevant point of comparison: it is equally fair to ask how UFMF stands against lightweight forecasters that already operate in real time. A like-for-like comparison on 3D occupancy metrics is not possible, because FlashOcc~\cite{yu2023flashocc} is a single-frame 3D perception model with no forecasting head, while FIERY~\cite{hu2021fiery} and PowerBEV~\cite{li2023powerbev} are 2D BEV forecasters that predict only the vehicle class and carry no height dimension. On their native metric, vehicle IoU, UFMF reaches 57.8 at 122 FPS and exceeds both by a large margin in accuracy and throughput simultaneously, even though it is evaluated at the harder 3\,s horizon rather than their 2\,s. The full comparison is tabulated in \Cref{apdx:2d_baselines}.

\subsection{Qualitative analysis}
\begin{figure*}[t]
\centering
\includegraphics[width=\linewidth]{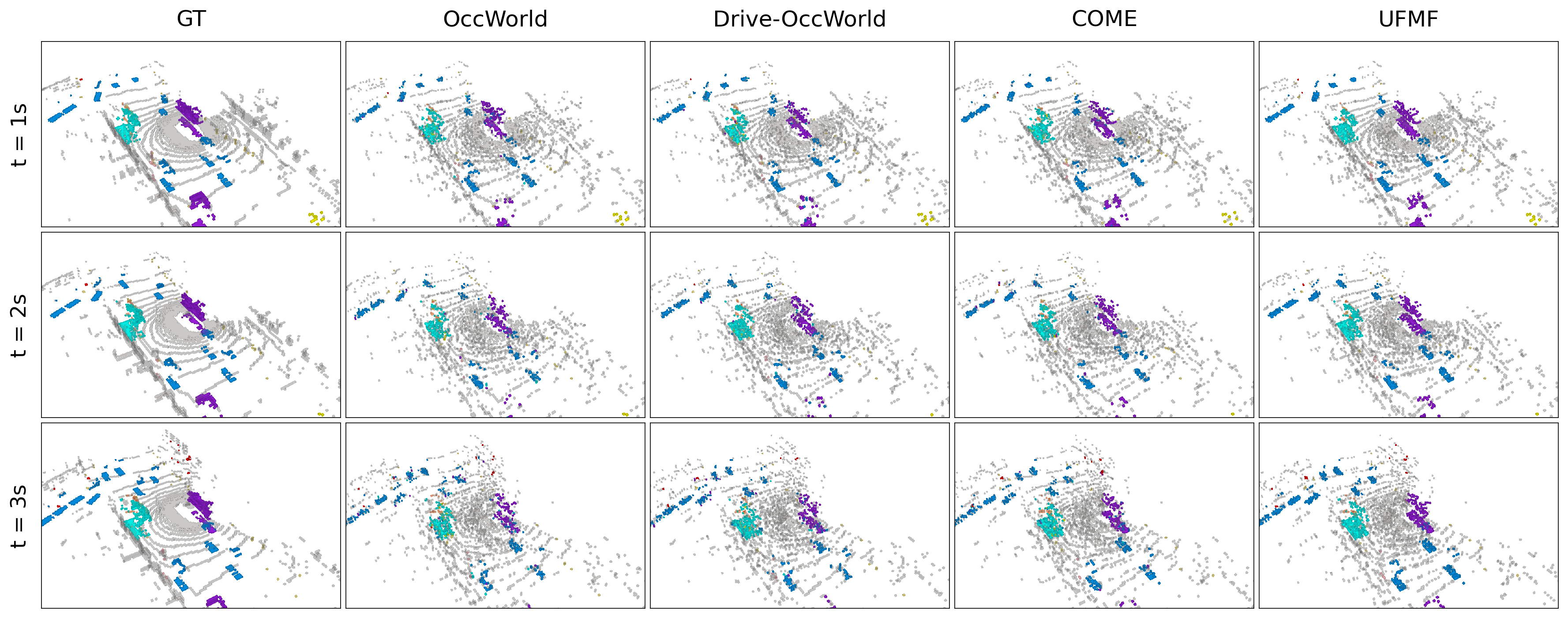}
\caption{3D qualitative comparison on one representative nuScenes~\cite{caesar2020nuscenes} validation scene over time. Columns show Ground Truth(GT), OccWorld, Drive-OccWorld, COME, and UFMF, while rows correspond to $t{=}1$ s, $t{=}2$ s, and $t{=}3$ s from the same oblique camera viewpoint. UFMF preserves cleaner actor extent and more stable scene structure as the horizon increases, especially around elongated vehicles and sparse dynamic clusters.}
\label{fig:qual_general}
\end{figure*}

\begin{figure}[t]
\centering
\includegraphics[width=\linewidth]{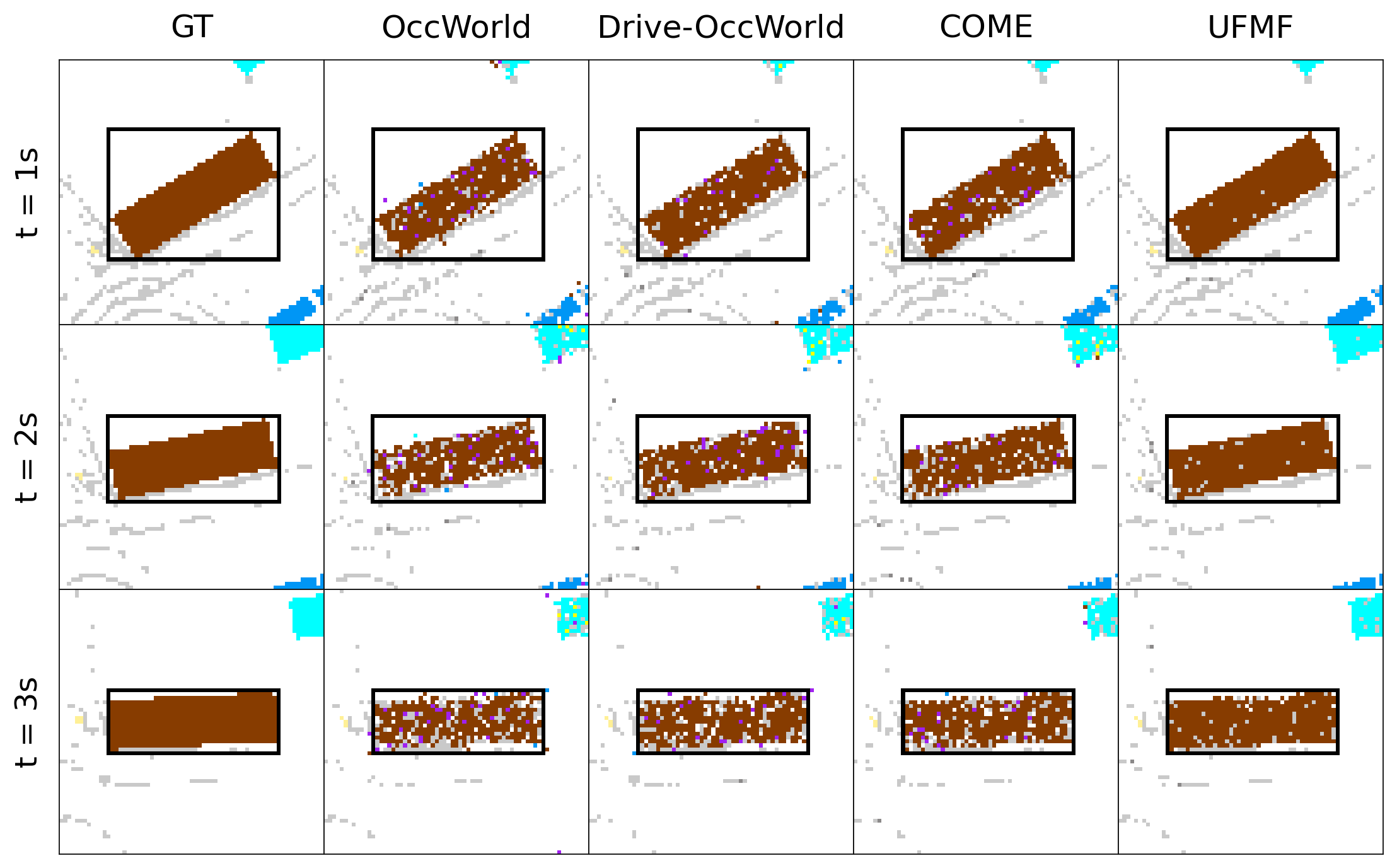}
\caption{BEV comparison on one validation scene focused on a single tracked actor from the fast-displacement bucket over time. Columns show GT, OccWorld, Drive-OccWorld, COME, and UFMF, while rows correspond to $t{=}1$ s, $t{=}2$ s, and $t{=}3$ s. Each subfigure is centered on the same tracked object, and the black box marks the region of interest. UFMF better preserves compact support and temporal alignment, whereas the weaker rows show stronger fragmentation, shrinkage, or stray occupancy.}
\label{fig:qual_fast}
\end{figure}

\Cref{fig:qual_general,fig:qual_fast} visualize the same question from two complementary viewpoints: whether a forecast remains structurally stable as the horizon increases. \Cref{fig:qual_general} shows a representative validation scene in 3D at $t{=}1$ s, $t{=}2$ s, and $t{=}3$ s. As the horizon increases, OccWorld and Drive-OccWorld progressively lose object extent or distort surrounding structure, while COME remains cleaner but still softens thin geometry and elongated actors. UFMF preserves sharper actor footprints and a more stable scene layout across time.

\Cref{fig:qual_fast} focuses on one large target actor from the fast-displacement bucket and keeps the same tracked object centered in all three horizons. This view makes the dynamic failure mode more explicit: the weaker rows either fragment the target, collapse it toward the centroid, or leak stray occupancy near the box boundary, whereas UFMF maintains compact support and better temporal alignment. Together, the 3D and BEV views show that the gain is not merely cosmetic at one endpoint, but persists over the whole short-horizon rollout.

\subsection{Ablation studies}
We structure the ablations to isolate kinematic and geometric structure rather than network scaling. The lightweight BEV baseline, which is also the comparison shown in \Cref{fig:teaser}, uses the same semantic-radar BEV encoder, diagonal state-space rollout, and channel-to-height lift as the full model, but removes the radar-guided VADP branch and the auxiliary world-frame and class-height constraints. Starting from this baseline, we first add VADP, then $\mathcal{L}_{\text{wf}}$ alone, then $\mathcal{L}_{\text{wf}}$ together with the trajectory loss $\mathcal{L}_{\text{traj}}$, and finally the class-conditioned height regularizer. The last row therefore differs from the preceding row only by adding $\mathcal{L}_{\text{height}}$.

\Cref{tab:ablation} shows a consistent progression. VADP alone raises fast-displacement mIoU from 0.298 to 0.403, confirming that explicit dynamic extrapolation addresses the dominant failure mode. Adding $\mathcal{L}_{\text{wf}}$ alone contributes a further gain to 0.418 fast mIoU, demonstrating that the world-frame temporal-difference and edge terms improve dynamic coherence even without the centroid trajectory constraint. Introducing $\mathcal{L}_{\text{traj}}$ on top raises the fast bucket to 0.435, which shows that the zone-level centroid loss provides complementary inter-horizon trajectory regularization. Finally, the class-conditioned height loss produces the full 0.542 overall mIoU and 0.45 fast-displacement mIoU. All configurations from +VADP onward share the same inference architecture and run at 122 FPS. Overall, the complete model improves the fast bucket by 15.2 points over the lightweight BEV baseline while remaining comfortably in the real-time regime.

\begin{table}[t]
\centering
\scriptsize
\setlength{\tabcolsep}{1.5pt}
\caption{Ablation study on the proposed components. VADP mainly improves the fast-displacement bucket, world-frame constraints stabilize slow- and fast-displacement trajectories, and the height regularizer preserves the final overall mIoU without sacrificing runtime.}
\begin{tabular*}{\columnwidth}{@{\extracolsep{\fill}}l c c c c c}
\toprule
Variant & mIoU(all) $\uparrow$ & Static $\uparrow$ & Slow $\uparrow$ & Fast $\uparrow$ & FPS $\uparrow$ / ms $\downarrow$ \\
\midrule
Lightweight BEV baseline & 0.433 & 0.521 & 0.426 & 0.298 & 138 / 7 \\
+ VADP & 0.529 & 0.618 & 0.524 & 0.403 & 122 / 8 \\
+ VADP + $\mathcal{L}_{\text{wf}}$ & 0.536 & 0.626 & 0.531 & 0.418 & 122 / 8 \\
+ VADP + $\mathcal{L}_{\text{wf}}$ + $\mathcal{L}_{\text{traj}}$ & 0.539 & 0.631 & 0.538 & 0.435 & 122 / 8 \\
+ $\mathcal{L}_{\text{height}}$ \textit{(Full UFMF)} & \textbf{0.542} & \textbf{0.632} & \textbf{0.541} & \textbf{0.450} & \textbf{122 / 8} \\
\bottomrule
\end{tabular*}
\label{tab:ablation}
\end{table}

\subsection{Robustness analysis}
\label{sec:robustness}

\paragraph{Radar degradation.}
Because VADP is conditioned on radar Doppler cues, it matters how much of UFMF's advantage survives when that modality is unavailable or unreliable. We evaluate three degraded conditions: radar removed entirely (zero input), radar replaced by semantic-delta-only VADP (no Doppler input to $g_{\text{vel}}$), and radar corrupted by additive Gaussian noise ($\sigma{=}5$\,m/s) on the Doppler channel. As \Cref{tab:radar_degrade} shows, removing radar altogether lowers fast-displacement mIoU to 0.407, which is still 10.9 points above the lightweight BEV baseline (0.298) and 7.1 points above OccWorld (0.336), while noisy radar costs only 1.9 points. UFMF therefore degrades gracefully rather than failing when the sensor is compromised.

\begin{table}[t]
\centering
\scriptsize
\caption{Radar modality isolation. The semantic-delta-only row removes radar from VADP; the radar-only row removes semantic deltas. Both retain the structural VADP mechanism, confirming that the architecture contributes independently of either input modality.}
\begin{tabular*}{\columnwidth}{@{\extracolsep{\fill}}l c c c}
\toprule
Radar condition & mIoU(all) $\uparrow$ & Fast $\uparrow$ & FPS $\uparrow$ / ms $\downarrow$ \\
\midrule
Full VADP (UFMF) & \textbf{0.542} & \textbf{0.450} & 122/8 \\
Noisy radar ($\sigma{=}5$\,m/s) & 0.534 & 0.431 & 122/8 \\
Radar-only VADP (no sem.\ delta) & 0.535 & 0.420 & 122/8 \\
Sem.-delta-only VADP (No radar) & 0.529 & 0.407 & 128/8 \\
\midrule
Lightweight BEV baseline & 0.433 & 0.298 & 138/7 \\
\bottomrule
\end{tabular*}
\label{tab:radar_degrade}
\end{table}

\paragraph{Predicted history inputs.}
UFMF also does not depend on the idealized ground-truth histories used by the standard protocol: when driven by BEV histories from a pre-trained BEVDet-Occ~\cite{huang2021bevdet} upstream model, it still exceeds COME with ground-truth history on the fast bucket. This experiment is reported in \Cref{apdx:predhistory}.

\subsection{Implementation details}
The model is trained with 6 history frames and 6 forecast horizons, a batch size of 4, learning rate $4\times10^{-5}$, weight decay $10^{-4}$, and 4 data-loading workers. Radar Doppler cues are clipped at 40 m/s before rasterization, and all targets are represented in the future-ego frame. All FPS numbers in \Cref{tab:main_results,tab:ablation} are measured on a single RTX 4090 using the same deployment setting used throughout the paper, and the same hardware and measurement protocol are used for every method and every ablation reported in this paper.

\section{Discussion}
\label{sec:discussion}

\paragraph{Radar dependency and modality fairness.}
VADP relies on radar Doppler cues, but \Cref{tab:radar_degrade} shows the model degrades gracefully: without radar, fast-displacement mIoU drops to 0.407---still exceeding COME (0.402) and OccWorld (0.336) while running 60$\times$ faster. The radar-only row confirms that Doppler cues alone also help, but the full system benefits from both modalities. The contribution is a system-level trade-off: under a real-time budget, a sub-0.5\,ms radar prior buys more dynamic fidelity per unit of latency than scaling a 3D transformer.

\paragraph{Structure vs.\ scale.}
The ablation trajectory (\Cref{tab:ablation}) shows that the dominant bottleneck in lightweight BEV forecasting is not temporal capacity but the absence of targeted kinematic and geometric constraints. Adding VADP, world-frame supervision, and height-profile regularization raises fast-displacement mIoU by 15.2 points without enlarging the backbone.

Extended discussions of sensor-grounded vs.\ dense-completion targets (e.g., Occ3D~\cite{tian2023occ3d}), scope of evaluation, and deployment considerations are provided in \Cref{apdx:datasetselection,apdx:discussion}.

\section{Conclusion}
We have presented UFMF, a real-time 4D occupancy forecaster that addresses the dynamic blank-collapse problem of lightweight BEV predictors without resorting to heavier generative world models. The key insight is that the dominant failure mode---loss of fast-moving actor support at longer horizons---can be resolved by injecting radar-guided kinematic priors and geometric training constraints into an otherwise cheap temporal core. On nuScenes, UFMF reaches 0.542 overall mIoU and 0.45 fast-displacement mIoU at 122 FPS, outperforming recent occupancy world models while running over 100$\times$ faster. Qualitative results confirm that these gains persist across the full 1--3\,s forecast window, producing cleaner support, less fragmentation, and more stable scene geometry over time.

\section*{LLM Usage Statement}
Claude (Anthropic) and GPT-4 (OpenAI) were used during the preparation of this paper. Their use was limited to the \textbf{writing} of the manuscript: polishing sentence structure, improving logical flow between paragraphs, correcting grammar, and suggesting concise phrasing for technical descriptions across all sections of the text. No generative AI tool was used for ideation, experimental design, code implementation, data analysis, figure generation, or the derivation of any mathematical formulation. All scientific claims, experimental results, architectural designs, and analyses are the sole work of the authors, who take full responsibility for the correctness and integrity of the entire manuscript.

{
    \small
    \bibliographystyle{ieeenat_fullname}
    \bibliography{main}
}

\clearpage
\appendix
\renewcommand{\thetable}{\Alph{section}\arabic{table}}

\section{Comparison with Efficient BEV Forecasters}
\label{apdx:2d_baselines}
\setcounter{table}{0}

Occupancy world models are not the only relevant point of comparison: it is equally fair to ask how UFMF stands against lightweight forecasters that already operate in real time. A like-for-like comparison on 3D occupancy metrics is not possible, because FlashOcc~\cite{yu2023flashocc} is a single-frame 3D perception model with no forecasting head, while FIERY~\cite{hu2021fiery} and PowerBEV~\cite{li2023powerbev} are 2D BEV forecasters that predict only the vehicle class and carry no height dimension. We therefore compare on their native metric, vehicle IoU (\Cref{tab:2d_baselines}). FIERY and PowerBEV forecast 2\,s, whereas UFMF is evaluated on the harder 3\,s horizon; even so, UFMF reaches 57.8 vehicle IoU at 122 FPS and exceeds both methods by a large margin in accuracy and throughput simultaneously.

\begin{table}[h]
\centering
\scriptsize
\caption{2D vehicle IoU against efficient BEV forecasters on nuScenes. UFMF is evaluated at the harder 3\,s horizon.}
\vspace{2mm}
\begin{tabular*}{\columnwidth}{@{\extracolsep{\fill}}l c c c c}
\toprule
Method & Veh.~IoU $\uparrow$ & Horizon & FPS $\uparrow$ & Dim. \\
\midrule
FIERY~\cite{hu2021fiery} & 36.7 & 2\,s & 28 & 2D \\
PowerBEV~\cite{li2023powerbev} & 39.3 & 2\,s & 52 & 2D \\
UFMF (veh.~class) & \textbf{57.8} & 3\,s & \textbf{122} & 3D$\to$2D \\
\bottomrule
\end{tabular*}
\label{tab:2d_baselines}
\end{table}

\section{Robustness to Predicted History Inputs}
\label{apdx:predhistory}
\setcounter{table}{0}

\Cref{tab:main_results} follows the standard 4D occupancy forecasting protocol, in which ground-truth BEV histories are supplied to every method so that forecasting quality is isolated from upstream perception error. To verify that UFMF does not depend on this idealized input, we additionally drive it with BEV histories produced by a pre-trained BEVDet-Occ~\cite{huang2021bevdet} upstream model. Accuracy drops modestly, from 0.542 to 0.521 mIoU(all) and from 0.450 to 0.427 on the fast-displacement bucket (\Cref{tab:pred_history}), yet UFMF with predicted history still exceeds COME with ground-truth history (0.402 fast) by 2.5 points. The forecaster is therefore robust to realistic upstream perception noise.

\begin{table}[h]
\centering
\scriptsize
\caption{Robustness to predicted (non-ground-truth) history inputs.}
\vspace{2mm}
\begin{tabular*}{\columnwidth}{@{\extracolsep{\fill}}l c c c}
\toprule
History source & mIoU(all) $\uparrow$ & Fast $\uparrow$ & FPS $\uparrow$ \\
\midrule
GT history (standard) & 0.542 & 0.450 & 122 \\
Predicted (BEVDet-Occ) & 0.521 & 0.427 & 122 \\
\bottomrule
\end{tabular*}
\label{tab:pred_history}
\end{table}

\section{Dataset Selection: Sensor-Grounded Forecasting vs.\ Scene Completion}
\label{apdx:datasetselection}

Recent 3D perception works often evaluate on dense completion benchmarks such as Occ3D~\cite{tian2023occ3d}, where the model is asked to infer occupancy in heavily occluded or genuinely unobservable regions, for example behind walls or underneath vehicles. That target is appropriate for scene completion, but it is not the target we want for a real-time future forecaster.

For UFMF, we instead train and evaluate on voxel grids derived directly from nuScenes~\cite{caesar2020nuscenes} LiDAR sweeps. This choice avoids a major source of optimization ambiguity. If the target requires the network to infer dense geometry in regions that were never physically observed, the model must spend capacity learning artificial completion patterns instead of the kinematic evolution of measurable dynamic actors. The resulting supervision is denser, but it is also less faithful to the forecasting problem we actually care about.

The architectural consequence is equally important. Dense completion targets tend to reward heavier 3D memory, volumetric refinement, or generative reconstruction modules, whereas our objective is reliable forecasting at 122 FPS. By restricting the target to physically supported occupancy, we train UFMF to extrapolate the future spatiotemporal evolution of observable structure rather than to fill in arbitrary hidden space. This choice is better aligned with downstream motion planning, where low latency and physically grounded predictions are prioritized over complete volumetric reconstruction.

\section{Extended Discussion}
\label{apdx:discussion}

\paragraph{Scope of evaluation.}
Our study is confined to nuScenes, which is at present the only public 2\,Hz benchmark providing synchronized camera, LiDAR and radar, and the benchmark on which all of our baselines are evaluated. Validating the same kinematic priors across additional sensor suites remains an open direction.

\paragraph{Deployment considerations.}
The 122 FPS figure measures the forecasting module alone. A complete perception-and-forecasting stack would add the cost of the upstream BEV encoder (e.g., BEVDet-Occ). We report the predicted-history experiment (\Cref{tab:pred_history}) to show that UFMF is robust to upstream noise, and note that the forecaster's 8\,ms latency leaves substantial headroom within a typical 50--100\,ms planning cycle.

\end{document}